\documentclass[conference]{IEEEtran}
\usepackage{amsmath,amssymb,amsfonts,amsthm}
\usepackage{graphicx}
\usepackage{textcomp}
\usepackage{xcolor}
\usepackage{hyperref}
\usepackage{booktabs}
\usepackage{subcaption}

\begin{document}

\title{How VLAs Fail Differently: Black-Box Action Monitoring\\Reveals Architecture-Specific Failure Signatures}

\author{\IEEEauthorblockN{Krishnam Gupta}
\IEEEauthorblockA{Independent Research \\
San Francisco, CA \\
kayjee1994@gmail.com}}

\maketitle

\begin{abstract}
We discover that VLA architectures fail in fundamentally different, predictable ways at the motor-command level.
Running VQ-BeT, Diffusion Policy, and ACT on identical evaluation protocols ($n{=}450$ episodes across PushT and ALOHA 14-DOF bimanual manipulation), we find:
(1)~direction reversal rate is a universal failure predictor across all three architectures (AUROC${=}$0.93, 0.79, 0.91; $p < 0.001$);
(2)~jerk monitoring is predictive only for discrete-token architectures, following a discrete-to-continuous gradient (0.88, 0.69, 0.41);
(3)~velocity violations alone are non-predictive everywhere (AUROC 0.41--0.69), yet velocity checking is the most common safety mechanism in VLA deployment code;
and (4)~for continuous-family VLAs, velocity monitoring provides effectively zero predictive signal (AUROC${=}$0.52 on ACT, 0.41 on Diffusion), proving that architecture-matched monitor selection is essential.
These results quantify a monitoring consequence of the well-known discrete/continuous VLA distinction: the two families produce qualitatively different failure signatures that require different monitors.
No single monitor works universally; architecture-matched selection is required.
This finding was enabled by \texttt{SafeContract}, a training-free, black-box action monitoring toolkit with conformal calibration.\footnote{Code: \url{https://github.com/krishnam94/vla-edge}}
\end{abstract}


\section{Introduction}

VLA models~\cite{rt2,openvla,smolvla} predict robot actions end-to-end from vision and language.
But what happens to those actions between the model's output layer and the robot's motors is largely unstudied.
OpenVLA's deployment script sends raw outputs directly to motors with zero bounds checking~\cite{openvla};
pi0 streams actions via WebSocket with no validation~\cite{pi0};
LeRobot's~\cite{lerobot} \texttt{EEBoundsAndSafety} processor is imperative, non-composable, and easy to omit.
The implicit assumption is that a good model produces safe actions.

Nobody has systematically characterized what VLA actions actually look like at the motor level, whether different architectures produce different failure signatures, or which monitoring signals predict task failure.
We did.
Using black-box action monitoring across three architectures, two tasks, and 450 episodes, we report four findings:

\begin{enumerate}
    \item \textbf{Direction reversal rate is a universal failure predictor.} It achieves AUROC${=}$0.93 (VQ-BeT), 0.79 (Diffusion), and 0.91 (ACT on 14-DOF ALOHA), all with $p < 0.001$. No other monitor is this consistent.
    \item \textbf{Jerk monitoring follows a discrete-to-continuous gradient.} Jerk AUROC decreases from 0.88 (discrete-token VQ-BeT) to 0.69 (chunk-based ACT) to 0.41 (continuous Diffusion). This gradient reflects the action generation mechanism, not the task.
    \item \textbf{Velocity monitoring is non-predictive everywhere.} AUROC ranges from 0.41 to 0.69 across all architectures, yet velocity checking is the most common safety mechanism in VLA deployment code.
    \item \textbf{Velocity monitoring is blind to continuous-family failures.} For ACT on ALOHA, velocity violation AUROC is 0.52 (random chance); for Diffusion, 0.41 (below chance). Despite being the most common safety mechanism, velocity checking provides no useful failure signal for these architectures.
\end{enumerate}

These findings establish a \emph{two-family pattern}: VLA architectures partition into a discrete family (autoregressive, VQ-VAE) where jerk and reversal rate are the strongest predictors, and a continuous family (diffusion, flow matching, action chunking) where momentum coherence and reversal rate matter more.
The practical consequence: practitioners must select monitors based on their VLA's architecture family, not apply a one-size-fits-all approach.

All experiments use \texttt{SafeContract}, a training-free, black-box action monitoring toolkit with conformal calibration and CUSUM shift detection~\cite{vovk2005,page1954}.
SafeContract requires no model access, no retraining, and adds $<$13\,$\mu$s per inference step.

\section{Method}
\label{sec:method}

\texttt{SafeContract} monitors and enforces constraints on VLA action outputs without model access.
We describe only the components needed to understand the experimental results; the full implementation is open-source.

\textbf{Safety contracts.}
A contract $C = (\mathbf{l}, \mathbf{u}, v_{\max})$ specifies per-joint bounds and a velocity limit.
Enforcement clips to bounds, clamps velocity, then re-clips (the re-clip is necessary because velocity clamping can push actions outside bounds).
All violations are logged with timestep, dimension, and magnitude.

\textbf{Conformal calibration.}
Hand-tuned thresholds are unreliable: a common choice of $v_{\max}{=}0.05$ clips a significant fraction of \emph{expert} actions.
We use split-conformal prediction~\cite{vovk2005}: demonstration episodes are split 80/20, nonconformity scores are computed on the calibration set, and the $(1{-}\alpha)(1 + 1/n)$ quantile yields bounds with a finite-sample coverage guarantee.
At $\alpha{=}0.05$, conformal bounds achieve 97.9\% holdout coverage while being 25\% tighter than the 4$\sigma$ heuristic.
Velocity limits are set per-joint at the 99th percentile of consecutive action deltas.

\textbf{Architecture-specific monitors.}
Beyond bounds and velocity, we compute five per-episode health metrics: (1)~\emph{direction reversal rate}: fraction of (timestep, joint) pairs where the action changes sign; (2)~\emph{jerk RMS}: root-mean-square of the third derivative of the action trajectory; (3)~\emph{momentum coherence}: cosine similarity between consecutive action deltas, measuring trajectory smoothness; (4)~\emph{spectral energy ratio}: fraction of signal energy in low frequencies; and (5)~\emph{stall detection}: consecutive steps with displacement below threshold $\tau$.
Each metric is computed per-episode and correlated with task success via AUROC.

\textbf{CUSUM shift detection.}
Conformal p-values feed a CUSUM detector~\cite{page1954}: $S_t = \max(0, S_{t-1} + \mathbf{1}[p_t < \alpha] - \alpha)$.
An alarm fires when $S_t > h$, with formal false-alarm bounds that ad-hoc EWMA cannot provide.

\section{Related Work}

\textbf{Training-time safety.}
SafeVLA~\cite{safevla} and RobustVLA~\cite{robustvla} improve safety via retraining or fine-tuning, complementary to runtime monitoring.

\textbf{Inference-time safety.}
AEGIS~\cite{aegis} solves a CBF-QP per step (requires dynamics models); SafeDiffuser~\cite{safediffuser} and CoDiG~\cite{codig} embed barriers in diffusion (architecture-specific); SafeDec~\cite{safedec} constrains decoding; ATACOM~\cite{atacom} projects onto constraint manifolds.
All require dynamics models, architecture access, or retraining.

\textbf{Failure detection.}
SAFE~\cite{safe2025} trains a failure detector on VLA internal features; FIPER~\cite{fiper2025} trains an RND network; Sentinel~\cite{sentinel} monitors temporal consistency.
These require training auxiliary models.
Kim et al.~\cite{kim2026guardrails} argue that modular guardrails are necessary for foundation-model robots.
While the discrete/continuous VLA distinction is well-known~\cite{rt2,diffusionpolicy}, no prior work has empirically compared monitor effectiveness across these families on the same task, or shown which monitors predict failure for which architectures.

\textbf{Conformal prediction for safety.}
Recent work validates conformal prediction for robot safety~\cite{safe2025,feldman2025}.
We use split-conformal bounds for action-space coverage and conformal p-values with CUSUM for shift detection.

\section{Experiments}

Platform: NVIDIA A40 GPU (RunPod), Python 3.12, LeRobot~\cite{lerobot} pretrained checkpoints.
Monitoring overhead: $<$0.001\% of VLA inference time (${\sim}$13\,$\mu$s per call).

\subsection{Same-Dataset Cross-Architecture Comparison}
\label{sec:crossarch}

We evaluate two architectures on the \emph{same} PushT dataset with identical seeds (0--199), contract (bounds $[0, 512]$, $v_{\max}{=}30$\,px/step), and success criterion (coverage ${\ge}0.95$).
This controls for task, environment, and evaluation protocol, isolating architecture effects.

\textbf{Diffusion Policy}~\cite{diffusionpolicy} (262M params, DDPM, $n{=}200$ per condition).
Without SafeContract: 58\% success (116/200, CI$_{95}$: [0.51, 0.65]).
With SafeContract: 57\% (114/200, CI$_{95}$: [0.50, 0.64]).
Fisher's $p = 0.92$.
SafeContract caught 772 velocity violations with no task degradation.

\textbf{VQ-BeT}~\cite{vqbet} (37.5M params, VQ-VAE + Behavior Transformer, $n{=}200$).
Without SafeContract: 56.5\% (113/200, CI$_{95}$: [0.50, 0.63]).
With SafeContract: 54.5\% (109/200, CI$_{95}$: [0.48, 0.61]).
Fisher's $p = 0.76$.
SafeContract caught 1{,}847 velocity violations with no significant degradation.

The architecture contrast: on the same task with identical data, VQ-BeT produces \textbf{2.4$\times$ more velocity violations} than Diffusion Policy (1{,}847 vs.\ 772).
This reflects a fundamental difference between discrete token transitions and smooth DDPM denoising.
VQ-BeT's mean jerk RMS is 2.7$\times$ higher (21.6 vs.\ 8.0), and it produces nearly zero stall steps (1 total) versus 117 for Diffusion.
The architectures fail in qualitatively different ways: VQ-BeT fails with jerky, oscillatory motion; Diffusion fails with smooth but stalled trajectories.

\textbf{Note on success rates.}
Our Diffusion Policy achieves 58\% vs.\ the published 87.4\%.
This gap reflects evaluation protocol: we use the LeRobot pretrained checkpoint with 300-step episodes and strict coverage $\ge 0.95$, while Chi et al.\ use 1{,}000-step episodes with relaxed thresholds.
VQ-BeT achieves 56.5\% (published: 57\%), confirming correct reproduction.
The cross-architecture comparison is valid because both models use identical evaluation conditions.

\subsection{Generalization: ACT on ALOHA 14-DOF}
\label{sec:aloha}

We extend to a fundamentally different setting: ACT~\cite{act} on ALOHA bimanual manipulation (14 joints, cube transfer task, $n{=}50$ per condition).

Without SafeContract: 62\% success (31/50, CI$_{95}$: [0.48, 0.74]).
With SafeContract: 56\% (28/50, CI$_{95}$: [0.42, 0.69]).
Fisher's $p = 0.68$.
SafeContract caught 4{,}758 velocity violations with no significant degradation.

\textbf{Velocity monitoring is blind to ACT's failures.}
Despite producing velocity violations when monitored (mean 37 per episode against conformal bounds), these violations carry \emph{no predictive signal}: velocity AUROC is 0.52 (random chance).
ACT fails 38\% of the time (19/50 episodes), but the failures are not caused by unsafe actions - they are caused by \emph{wrong} actions that happen to look physically valid.
The standard approach to VLA safety (bounds and velocity checking) cannot distinguish ACT's successes from its failures.
Detecting these failures requires monitors that capture action \emph{quality} (reversal rate: AUROC${=}$0.91), not just action magnitude.

\subsection{Failure Prediction: Which Monitors Work?}
\label{sec:failure_prediction}

\begin{figure}[t]
    \centering
    \includegraphics[width=\columnwidth]{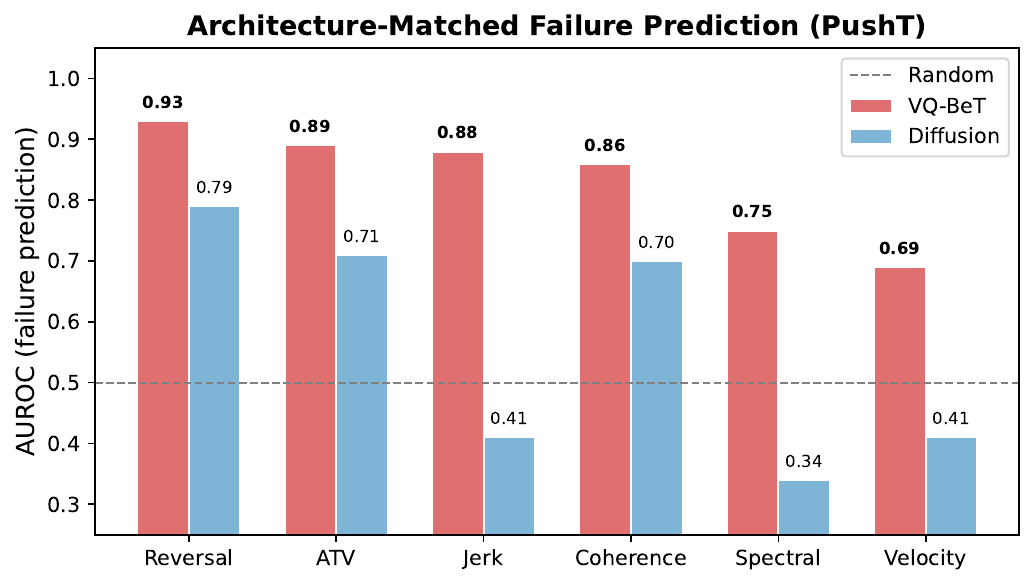}
    \caption{Architecture-matched failure prediction (PushT, $n{=}100$). Reversal rate predicts failure across both architectures (0.93 and 0.79). Jerk is predictive only for discrete-token VQ-BeT (0.88 vs.\ 0.41). No single monitor is universally best.}
    \label{fig:auroc}
\end{figure}

\begin{figure}[t]
    \centering
    \includegraphics[width=\columnwidth]{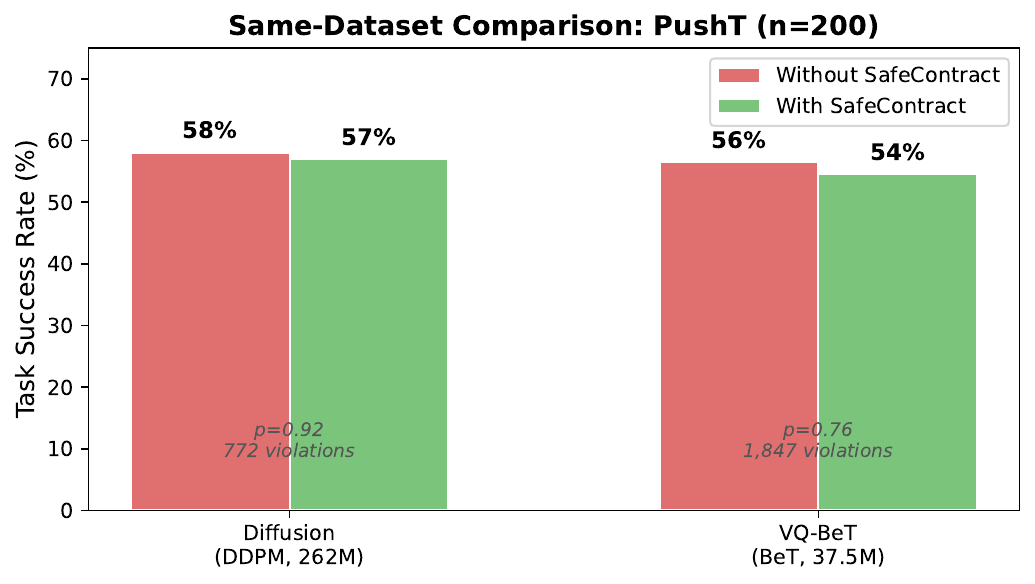}
    \caption{Same-dataset closed-loop comparison on PushT ($n{=}200$ per condition). VQ-BeT produces 2.4$\times$ more violations than Diffusion Policy (1{,}847 vs.\ 772) with qualitatively different failure signatures.}
    \label{fig:closedloop}
\end{figure}

We compute per-episode health metrics on a diagnostic subset ($n{=}100$ for PushT, $n{=}50$ for ALOHA) and measure AUROC for predicting task failure.
Table~\ref{tab:crossarch} and Figs.~\ref{fig:auroc}--\ref{fig:closedloop} report all results; the key findings are:

\textbf{Direction reversal rate is the universal winner.}
It achieves the highest AUROC on all three architectures: 0.93 on VQ-BeT, 0.79 on Diffusion, and 0.91 on ACT/ALOHA ($p < 0.001$ all three).
This is the only monitor that works across both families and across 2D and 14-DOF action spaces.
A high reversal rate indicates the policy is oscillating rather than progressing - an architecture-agnostic signature of indecision.

\textbf{Jerk follows a discrete-to-continuous gradient.}
Jerk AUROC: 0.88 (VQ-BeT, discrete tokens), 0.69 (ACT, continuous chunks with boundary effects), 0.41 (Diffusion, continuous denoising).
This gradient is not task-dependent; it reflects the action generation mechanism.
Discrete-token models produce quantization jumps between codebook entries that manifest as high jerk.
Diffusion's iterative denoising smooths these away.
ACT sits between: smooth within chunks, but chunk boundaries introduce moderate jerk.

\textbf{Velocity violations are non-predictive.}
AUROC: 0.69 (VQ-BeT), 0.52 (ACT), 0.41 (Diffusion).
For Diffusion, velocity AUROC is below chance (0.41), meaning episodes with \emph{more} violations are slightly \emph{more} likely to succeed.
This is because successful episodes involve larger, more confident motions.
Velocity violation counting, despite being the default safety mechanism, provides no useful failure signal.

\textbf{Per-family monitor profiles.}
The discrete family (VQ-BeT) shows high AUROC across nearly all monitors: jerk violations (0.89), total variation (0.89), momentum coherence (0.86), spectral energy (0.75).
The continuous family (Diffusion, ACT) shows high reversal rate and moderate momentum coherence, but jerk and spectral monitors are at or below chance.
Stall detection is non-predictive for all three architectures in our experiments (AUROC $<$ 0.53), though we note the PushT task may not induce stall-type failures.

\begin{table*}[t]
\centering
\caption{Cross-architecture comparison: deployment statistics and failure prediction AUROCs. PushT: $n{=}200$ per condition (same seeds/contract), AUROCs on $n{=}100$. ALOHA: $n{=}50$ per condition, AUROCs on $n{=}50$. Bold: highest AUROC per monitor.}
\label{tab:crossarch}
\small
\begin{tabular}{lccc}
\toprule
 & \textbf{Diffusion Policy (PushT)} & \textbf{VQ-BeT (PushT)} & \textbf{ACT (ALOHA 14-DOF)} \\
\midrule
Architecture family & Continuous & Discrete & Continuous \\
Params & 262M & 37.5M & 51.6M \\
Success (no / with contract) & 58\% / 57\% & 56.5\% / 54.5\% & 62\% / 56\% \\
Fisher's $p$ & 0.92 & 0.76 & 0.68 \\
Total violations & 772 & \textbf{1{,}847} & 4{,}758 \\
Mean jerk RMS & 8.0 & \textbf{21.6} & 0.054 \\
Total stall steps & 117 & \textbf{1} & -- \\
\midrule
\multicolumn{4}{l}{\textit{Failure Prediction AUROCs (higher = more predictive)}} \\
\midrule
Reversal rate & 0.79 {\scriptsize[.69,.88]} & \textbf{0.93} {\scriptsize[.88,.97]} & \textbf{0.91} {\scriptsize[.79,.99]} \\
Jerk (mean) & 0.41 {\scriptsize[.30,.54]} & \textbf{0.88} {\scriptsize[.81,.94]} & 0.69 {\scriptsize[.54,.84]} \\
Momentum coherence & 0.70 & \textbf{0.86} & 0.71 \\
Spectral energy ratio & 0.34 & \textbf{0.75} & 0.74 \\
Total variation & 0.71 & \textbf{0.89} & 0.75 \\
Velocity violations & 0.41 & 0.69 & 0.52 \\
Stall rate & 0.50 & 0.49 & 0.53 \\
\midrule
\multicolumn{4}{l}{\textit{Recommended Monitors}} \\
\midrule
Primary & reversal rate & reversal rate + jerk & reversal rate \\
Secondary & momentum coherence & momentum + spectral & momentum + spectral \\
Avoid & jerk, velocity, spectral & stall & velocity, stall \\
\bottomrule
\end{tabular}
\end{table*}

\subsection{Enforcement Without Degradation}

Across all three architectures and 450 total episodes, SafeContract monitors and enforces without statistically significant task degradation: Fisher's $p{=}0.92$ (Diffusion), $p{=}0.76$ (VQ-BeT), and $p{=}0.68$ (ACT).
Conformal calibration is key: data-driven bounds are tight enough to be informative but loose enough to avoid clipping correct actions.
The with-contract violations (772, 1{,}847, 4{,}758) come from conformally-calibrated limits, meaning they represent genuinely anomalous actions, not overly conservative thresholds.

\section{Discussion}

\textbf{The two-family pattern.}
Our results reveal a consistent pattern: VLA architectures partition into two families with distinct failure signatures.
The \emph{discrete family} (autoregressive, VQ-VAE) produces quantization artifacts that manifest as high jerk, high reversal rates, and velocity spikes.
Jerk and reversal rate are strongly predictive (AUROC $>$ 0.85).
The \emph{continuous family} (diffusion, flow matching, action chunking) produces smooth trajectories where the primary failure mode is not constraint violation but behavioral error - wrong actions that look physically valid.
Reversal rate remains predictive (0.79--0.91), but jerk drops to chance (0.41 for pure diffusion).

ACT on ALOHA sits between the families: as a continuous model with chunk boundary effects, its jerk AUROC (0.69) is intermediate.
The gradient VQ-BeT (0.88) $\rightarrow$ ACT (0.69) $\rightarrow$ Diffusion (0.41) tracks the discrete-to-continuous spectrum, though we note this is partially confounded by task dimensionality (PushT 2D vs.\ ALOHA 14-DOF); isolating the architecture effect requires testing all three on the same task.

\textbf{The velocity monitoring blind spot.}
The ACT result is perhaps the most practically significant.
Despite producing velocity violations against conformal bounds, these violations carry no predictive signal (AUROC${=}$0.52, random chance).
ACT fails 38\% of the time, but the failures are behavioral, not physical: the robot executes actions that simply do not accomplish the task.
Detecting these failures requires monitors that capture action \emph{quality} (reversal rate: 0.91, momentum coherence: 0.71), not just action \emph{magnitude} (velocity: 0.52).
This is a strong negative result about the most common VLA safety mechanism.

\textbf{Practical recommendations.}
Given these findings, we propose architecture-matched monitoring:
For discrete-token models (VQ-BeT, OpenVLA): monitor jerk RMS and reversal rate.
For continuous denoisers (Diffusion Policy, flow-matching): monitor reversal rate and momentum coherence; skip jerk.
For action-chunked models (ACT): monitor reversal rate and watch for chunk boundary artifacts.
In all cases, do not rely on velocity violation counting as a primary failure signal.
\texttt{SafeContract}'s \texttt{SafetyGuard.from\_demos()} API calibrates and selects monitors automatically from demonstration data.

\textbf{Scope and limitations.}
(1)~All experiments are in simulation; real-robot patterns may differ.
(2)~Three of eleven known VLA architecture types~\cite{openvla,pi0} are tested; the discrete-vs-continuous framing likely generalizes but is unverified for autoregressive (OpenVLA) and flow-matching (pi0) architectures.
(3)~Emerging hybrid architectures (AR reasoning + diffusion actions) can produce valid trajectories toward wrong targets, invisible to any action-space monitor.
Semantic correctness is a property of the (action, observation, instruction) triple, not of the action alone; detecting it requires perception~\cite{fiper2025} and task-level~\cite{sentinel} monitoring.
(4)~SafeContract is monitoring infrastructure, not a safety guarantee. It composes with CBF-based constraints~\cite{aegis} as the innermost layer of a defense-in-depth stack.

\textbf{Future work.}
Extending to autoregressive (OpenVLA) and flow-matching (pi0) architectures would validate the two-family pattern across a broader set.
Adaptive conformal inference~\cite{gibbs2021} would maintain coverage guarantees during distribution shift.
Real-robot validation is essential before deployment recommendations can be made with confidence.
With the EU AI Act (Article 9) mandating continuous monitoring for high-risk AI systems by 2027, architecture-matched action monitoring is an urgent practical need.

\end{document}